\documentclass[11pt,a4paper]{article}
\usepackage[hyperref]{eacl2021}
\usepackage{times}
\usepackage{latexsym}
\usepackage{amsfonts}
\usepackage{xcolor}

\usepackage{graphicx}
\usepackage{subfigure}
\usepackage{tipa}
\usepackage{listings}
\definecolor{codegreen}{rgb}{0,0.6,0}
\definecolor{codegray}{rgb}{0.5,0.5,0.5}
\definecolor{codepurple}{rgb}{0.58,0,0.82}
\definecolor{backcolour}{rgb}{0.95,0.95,0.92}

\lstdefinestyle{mystyle}{
    backgroundcolor=\color{backcolour},   
    commentstyle=\color{codegreen},
    keywordstyle=\color{magenta},
    numberstyle=\tiny\color{codegray},
    stringstyle=\color{codepurple},
    basicstyle=\ttfamily,
    breakatwhitespace=false,         
    breaklines=true,                 
    captionpos=b,                    
    keepspaces=true,                 
    numbers=left,                    
    numbersep=5pt,                  
    showspaces=false,                
    showstringspaces=false,
    showtabs=false,                  
    tabsize=2
}

\usepackage{hyperref}
\hypersetup{
    colorlinks=true,
    linkcolor=blue,
    filecolor=magenta,      
    urlcolor=cyan,
}
\usepackage{microtype}
\aclfinalcopy 

\title{NLP for Ghanaian Languages}
\author{Paul Azunre$^{1*}$, Salomey Osei$^{2*}$, Salomey Afua Addo$^{3*}$,  Lawrence Asamoah Adu-Gyamfi$^{*}$, \\
\textbf{Stephen Moore$^{4*}$, Bernard Adabankah$^{5*}$, Bernard Opoku$^{6*}$, Clara Asare-Nyarko$^{4*}$} \\
\textbf{Samuel Nyarko$^{15*}$, Cynthia Amoaba$^{*}$, Esther Dansoa Appiah$^{8*}$, Felix Akwerh$^{2*}$} \\
\textbf{Richard Nii Lante Lawson$^{9*}$, Joel Budu$^{10*}$, Emmanuel Debrah$^{4*}$, Nana Boateng$^{1*}$} \\
\textbf{Wisdom Ofori$^{*}$, Edwin Buabeng-Munkoh$^{*}$, Franklin Adjei$^{11*}$, Isaac K. E. Ampomah$^{12*}$} \\
\textbf{Joseph Otoo$^{13*}$, Reindorf Borkor$^{2*}$, Standylove Birago Mensah$^{2*}$, Lucien Mensah$^{7*}$} \\
\textbf{Mark Amoako Marcel$^{*}$, Anokye Acheampong Amponsah$^{14*}$, and James Ben Hayfron-Acquah$^{2*}.$} \\ \\

 $^*$ NLP Ghana,$^1$ Algorine,$^2$ Kwame Nkrumah University of Science and Technology, \\
 $^3$ Leuphana University Luneburg,$^4$University of Cape Coast, 
 $^5$ Edinburgh Napier University, \\ $^6$ Accra Institute of Technology,$^7$ Tulane University, $^{8}$ University of Tromso, $^{9}$ AiMlCamp, \\ $^{10}$ University of Strathclyde, $^{11}$ Azubi Africa, $^{12}$ Ulster University, \\
 $^{13}$ Centre for Research, Data Science and IT Solutions, $^{14}$ University of Energy and Natural Resources, \\ $^{15}$ Integrated Geospatial Intelligence Application Centre, SRH Berlin University of Applied Science.
  }

\date{}
\begin{document}
\maketitle
\begin{abstract}
\vspace{-0.62mm} NLP Ghana is an open-source non-profit organisation aiming to advance the development and adoption of state-of-the-art NLP techniques and digital language tools to Ghanaian languages and problems. In this paper, we first present the motivation and necessity for the efforts of the organisation; by introducing some popular Ghanaian languages while presenting the state of NLP in Ghana. We then present the NLP Ghana organisation and outline its aims, scope of work, some of the methods employed and contributions made thus far in the NLP community in Ghana.
\end{abstract}

\section{Introduction}
The advancement in machine learning computational power coupled with the recent investment within the domain by technological companies has stimulated considerable interest and brought about a legion of applications in natural language digitisation in developed countries, with much focus on the English language \citep{martinus2019focus}. In fact, English is the most computerised language in the world and corpora in the language are best documented, due to availability of authentic electronic texts, and its usage as both mother tongue and second language \citep{leech1992computers, kenny2014lexis,martinus2019focus,varab2020danewsroom}.

This is not the case for Ghanaian languages and other languages across Africa, although the continent has over 2000 languages and the highest density of languages in the world \citep{tiayon2005community}. Many of these languages in Africa are yet to evolve from simple online existence to optimal online presence \citep{writingandtranslationinAfricanlanguages}. Though there seems to be progress in the development of applications as far as Ghanaian languages are concerned, their usability is limited, leaving much room for improvement in their operationalisation and adoption into the Ghanaian tonal, multilingual and digitisation systems 
\citep{martinus2019focus,kugler2016tone}.

In the much-applauded interventions by Google and Microsoft through their translation services, quite a number of  African languages have been integrated, but Ghanaian languages are excluded \citep{GoogleLanguageSupport,Microsofttranslate}. A historic move worth mentioning is Baidu Translate's incorporation of the Twi language in their translation service. Notwithstanding, its output in the Twi language semantically is questionable, as it is often does not make sense and truncates Twi words \citep{BaiduTranslate}. In fact, nothing can be more demotivating than situations where professional writers who work with African languages, students, tourists, among others, cannot use otherwise commonly available translation technology to perform simple tasks \citep{tiayon2005community}.
Major challenges boil down to lack of good (indeed often any) training data, as well as lack of adoption of major language technologies for local problems. This makes it difficult for the writing systems of many African languages to effectively pass the tests of user-friendliness and internet visibility \citep{ writingandtranslationinAfricanlanguages}.

\citet{Lackfunds} sufficiently underscores that not much funding is available for translation from and into African languages. Moreover, there has been general failure to use Ghanaian languages together with other African languages in various specialised fields. This has equally hindered their development in the areas of electronic and online resources, as well as human language technology \citep{shoba2018exploring}.

NLP Ghana seeks to close some of the gaps that were just identified. While the focus is on Ghanaian languages, the tools and techniques are developed with an additional goal of generalisability to other low resource language scenarios.

\section{State of NLP in Ghana}
Ghanaian languages are yet to evolve to optimal online presence and internet visibility. To date, there is no reliable machine translation system for any  Ghanaian language \citep{Googletranslate}. This makes it harder for the global Ghanaian diaspora to learn their own languages. Ghanaian languages and culture also risk not being preserved electronically in an increasingly digitised future. Consequently,  service providers and health workers trying to reach remote areas hit by emergencies, disasters, etc. face needless additional obstacles to providing life-saving care. Beyond translation, fundamental tools for computational analysis such as corpus-processing and analysis tools are lacking. Tools for summarisation, classification, language detection, voice-to-text transcription are limited, and in most cases completely non-existent \citep{varab2020danewsroom}.  This is a major risk to Ghanaian national security. Availability of these tools is directly correlated with the sophistication and efficiency of cyber-security solutions that can be deployed to defend critical social, cultural, and cyber infrastructure from both internal and external threats \citep{chambers2018detecting,siracusano2019poster}.

\section{Background Information on Ghanaian Languages}
Ghana is a multilingual country with at least 75 local languages \citep{Ethniclinguisticgroups}. Gur languages are spoken in the northern part by about 24\% of the total population while Kwa languages are spoken in the southern part by about 75\% of the population \citep{schneider2004handbook}. Research suggests that about 51\% of adults in Ghana are literate in both English and an indigenous language, while smaller portions of the population are literate in either English only or Ghanaian language only \citep{adika2012english}. There are nine (9) government-sponsored Ghanaian languages so far studied in Ghanaian educational institutions, namely, Akan (Akuapem Twi, Asante Twi and Fante dialects only), Dagaare, Dagbani, Dangme, Ewe, Ga, Gonja, Kasem, Mfantse and Nzema \citep{abokyi2018interface}.

Akan is the most commonly spoken Ghanaian language and the most used after English in the electronic public media. In some cases, it is used more than English \citep{browns,adika2012english}.
 Languages that belong to the same ethnic group are reciprocally understandable \citep{chen2014machine,noels2014language}. For instance, languages such as Dagbani and Mampelle, popularly spoken in the northern part of Ghana are mutually intelligible with the Frafra and Waali languages of the Upper Regions of Ghana. These four languages are of Mole-Dagbani ethnicity.  \citep{abokyi2018interface,LanguageandReligion}.
The chart in Table \ref{s1} shows language speaker data provided by Ethnologue \cite{campbell2008ethnologue}.

\begin{table}
\centering
\begin{tabular}{lr}
\hline \textbf{Languages} & \textbf{Number of Speakers} \\ \hline
Akan (Fante/Twi)  &  9,100,000\\
Ghanaian Pidgin English & 5,000,000\\
Ewe & 3,820,000 \\
Abron & 1,170,000 \\
Dagbani & 1,160,000 \\
Dangme & 1,020,000 \\
Dagaare & 924,000 \\
Konkomba & 831,000 \\
Ga & 745,000 \\
Farefare & 638,000 \\
Kusaal & 535,000 \\
Mampruli & 316,000 \\
Gonja & 310,000 \\
Sehwi & 305,000 \\
Nzema & 299,000 \\
Wasa & 273,000 \\
\hline
\end{tabular}
\caption{Common government sponsored languages in Ghana, \citep{wiki:xxx}}
\label{s1}
\end{table}
Ghanaian Pidgin is noteworthy in that it is not a government-sponsored language, and is an English Creole that is a variant of the West African Pidgin English. It is spoken heavily in the southern parts of the country and used heavily by the youth on social media and online in general \citep{deumert2020sub,suglo2015language}. \citep{schneider2004handbook} identifies two varieties which he describes as \textit{uneducated} and \textit{educated/student} varieties of Ghanaian Pidgin English. The former is usually used as lingua franca in highly multilingual contexts while the latter is often used as in-group language to express solidarity. The main differences between them are lexical rather than structural. NLP Ghana hopes to add Ghanaian Pidgin English to its projects based on the crucial role it plays in some of these Ghanaian communities. 

With respect to selecting languages for NLP Ghana projects in general, representative languages from both the southern and northern parts of the country are considered for full representation. Several Ghanaian languages are important not only for Ghana, but also its surrounding West African countries. Since enhanced regional trade is an important target societal benefit for the advances in language technologies we are discussing, this is also taken into account. For instance, Akan is spoken in C\^{o}te d’Ivoire while Ewe is spoken in Togo, Benin and Nigeria. Moreover, Gurune or Frafra is equally spoken in Burkina Faso \citep{deumert2020sub}.

Languages in the southern part of Ghana selected for initial exploration are Akan (Asante Twi,  Akuapem Twi and Fante dialects), Ewe, Ga  and Pidgin. Akan, Ewe and Pidgin are among the most widely-spoken languages in  Ghana, making  them  straight-forward additions to the NLP Ghana target languages \citep{deumert2020sub}. Ga is spoken as  native language in the capital Accra and may be particularly valuable to international travelers. This makes Ga another straight-forward addition to the initial target language set. 

Northern Ghanaian languages initially selected for NLP Ghana projects include Dagaare, Dagbani, Gonja and Gurune (Frafra). Northern Ghana is typically perceived as being poorer, with less local language digitisation, education and resources \citep{yaro2010contours}. This is sometimes attributed to the region being far away from the capital and therefore further away from the most lucrative economic activities. Including these languages in NLP Ghana's projects ensures better representation for equitable and sustainable development.  

\section{NLP Ghana Agenda}
NLP Ghana is an open-source movement of like-minded volunteers who have dedicated their skills and time to building an ecosystem of:
\begin{enumerate}
    \item Open-source data sets.
    \item Open-source computational methods.
    \item NLP researchers, scientists and practitioners excited about revolutionising and improving every aspect of Ghanaian life through this increasingly powerful and influential technology.
\end{enumerate}
 Although NLP Ghana is currently working on Ghanaian languages, particularly Akan due to number of speakers, the ultimate goal is to develop language tools applicable throughout the West African sub-region and beyond, complementing efforts such as \citep{nekoto2020participatory}.
 
NLP Ghana seeks to develop better data sources to train state-of-the-art (SOTA) NLP techniques for Ghanaian languages and to contribute to adapting SOTA techniques to work better in a lower resource setting. In other words, it aims to build functional systems for local applications such as a ``Google Translate  for the Ghanaian languages". The group equally seeks to train  and  benchmark  algorithms  for  a  number  of  crucial  tasks  in  these  languages --  translation,  named  entity  recognition  (NER),  POS  tagging  and sentiment  analysis, training classical text embeddings such as word2vec and FastText, as well as fine-tuning  contextual embeddings  such  as  BERT \citep{devlin2018bert}, DistilBERT \citep{sanh2019distilbert}  and  RoBERTA \citep{liu2019roberta}. 
NLP Ghana is therefore open to both local and international entity collaborations in the pursuit of its mission.

\section{NLP Ghana Contributions}

NLP Ghana has developed translators between English and some of the most-widely spoken Ghanaian languages -- Twi, Ewe and Ga. Our translators are already available to the general Ghanaian public as a \href{http://translate.ghananlp.org/}{Web Application}, as well as mobile applications via the \href{https://play.google.com/web/store/apps/details?id=com.nlpghana.khaya&hl=en&gl=US} {Google Play Store} and the \href{https://apps.apple.com/no/app/khaya/id1549294522}{Apple Store}. Response from the public has been largely positive, suggesting the crucial need for such services.

Embeddings have also been developed for Akan as the most widely spoken Ghanaian language \citep{ghananlp-ABENA}. These include both static embeddings such as fastText \citep{fastText} and contextual embeddings such as BERT \citep{devlin2018bert}. Both models have been open-sourced and made available to the public via a few lines of Python code.
 
As indicated earlier,  NLP Ghana also aims to produce large training data sets for Akan, Ewe, Ga and other Ghanaian languages -- starting with the first three since a functional translator has been developed for these languages. Work on training data for the Akuapem dialect of Akan was recently completed as part of a collaboration with \href{https://zindi.africa/}{Zindi Africa}. One of the goals of NLP Ghana is to create at least $50,000$  sentences  for each target language,  providing  a  reasonably  sized data set for fine-tuning modern neural network architectures on the data.

Data used to augment internally-created data for these projects include, but is not limited to, the JW300 multilingual data set \citep{agic2020jw300} and the Bible \citep{BibleCorp}.

\section{Participation and Methodology}
Our volunteers mainly identify as students (both graduate and undergraduate), ML researchers, data scientists, mathematicians, engineers, lecturers, programmers and local language instructors. At the moment, member count exceeds one hundred ($100$). The combined skills of members are utilised in various teams -- Data, Engineering, Research and Communications -- to shape and execute the broad NLP Ghana agenda. 

Specifically, the Data Team is responsible for data collection and storage while the Engineering Team manages data, networks and platforms. The Research Team leads the way directing technical agenda and devising strategies in an effort to optimise the execution of research programmes and meeting set Key Performance Indicators (KPIs). The Research Team is further divided into Unsupervised Methods, Supervised Methods, and Evaluation, by corresponding technical area. The Communications Team is also responsible for internal and external information flow. The team does this by liaising with stakeholders, interacting with product users and relaying feedback to teams to ensure continuous improvement of products.

Two main streams have been used to collect data: crowd-sourcing and human-correction of machine translated data from  in-house translator codes. The former is acquired by collecting voluntary responses of people through Google Form surveys. This exercise allows people to translate a set of randomly drawn English sentences into Akan. In total, about $697$ sentence pairs have been generated with this method. The latter generated about $50,000$ preliminary translations which were distributed to well-qualified native speakers to revise. This has yielded approximately $25,000$ translations into Akuapem Twi which is inclined towards Asante Twi in terms of tone and vocabulary \citep{ghananlp-DATA}.

The process of collecting data and processing them is not without challenges. One of the major challenges has been the inability to employ professional translators to verify and review machine translations due to financial constraints. Moving forward, NLP Ghana hopes to extend data collection capabilities beyond text data to other forms of data such as audio data (oral corpora) on several Ghanaian languages. The group also aims to annotate  data sets to enhance the works of NLP researchers in carrying out downstream NLP tasks such as Named Entity Recognition (NER), Part of Speech (POS) tagging etc. This effort will require significant funding by various stakeholders to yield a good quantity of quality data.

\section{Conclusion}
Research works on NLP have provided several indispensable tools useful in this modern internet age. This paper presented the state of NLP applications to Ghanaian languages such as Akan, Ga, and Ewe. One of the major challenges has been the lack of evaluation data sets to efficiently develop machine learning models for NLP tasks including machine translation, named entity recognition and document classification for Ghanaian languages. NLP Ghana has built an open-source community of researchers with different levels of expertise, working together to develop data sources, techniques and models for Ghanaian and other low-resource languages. To this end, the group has already open-sourced some models and data sets to further research activities for Ghanaian languages.

\section*{Acknowledgments}
We are grateful to the Microsoft for Startups Social Impact Program -- for supporting this research effort via providing GPU compute through Algorine Research. We would like to thank Julia Kreutzer, Jade Abbot and Emmanuel Agbeli for their constructive feedback. We are grateful to the reviewers for their valuable comments. We would also like to thank the \href{https://gajreport.com}{Ghanaian American Journal} for their work in sharing our vision with the Ghanaian public.

\bibliography{anthology,eacl2021}
\bibliographystyle{acl_natbib}
\end{document}